\documentclass[10pt,twocolumn,letterpaper]{article}

\usepackage{times}
\usepackage{epsfig}
\usepackage{microtype}
\usepackage{graphicx}
\usepackage{amsmath}
\usepackage{amssymb}
\usepackage{url}
\usepackage{wrapfig}
\usepackage{listings}
\usepackage{xcolor}
\usepackage{subcaption}
\usepackage{booktabs}
\usepackage{wacv}
\usepackage[bookmarks=false]{hyperref}

\colorlet{punct}{red!60!black}
\definecolor{background}{HTML}{EEEEEE}
\definecolor{delim}{RGB}{20,105,176}
\colorlet{numb}{magenta!60!black}
\lstdefinelanguage{json}{
    basicstyle=\tiny\ttfamily,
    numbers=left,
    numberstyle=\scriptsize,
    stepnumber=1,
    numbersep=6pt,
    showstringspaces=false,
    breaklines=true,
    frame=lines,
    backgroundcolor=\color{background},
    literate=
     *{0}{{{\color{numb}0}}}{1}
      {1}{{{\color{numb}1}}}{1}
      {2}{{{\color{numb}2}}}{1}
      {3}{{{\color{numb}3}}}{1}
      {4}{{{\color{numb}4}}}{1}
      {5}{{{\color{numb}5}}}{1}
      {6}{{{\color{numb}6}}}{1}
      {7}{{{\color{numb}7}}}{1}
      {8}{{{\color{numb}8}}}{1}
      {9}{{{\color{numb}9}}}{1}
      {:}{{{\color{punct}{:}}}}{1}
      {,}{{{\color{punct}{,}}}}{1}
      {\{}{{{\color{delim}{\{}}}}{1}
      {\}}{{{\color{delim}{\}}}}}{1}
      {[}{{{\color{delim}{[}}}}{1}
      {]}{{{\color{delim}{]}}}}{1},
}
\wacvfinalcopy % *** Uncomment this line for the final submission

\def\wacvPaperID{674} % *** Enter the wacv Paper ID here

\ifwacvfinal\fi
\begin{document}

%%%%%%%%% TITLE
\title{Attention Based Natural Language Grounding by Navigating Virtual Environment}

% Authors at the same institution
% \author{First Author \hspace{2cm} Second Author \\
% Authors at different institutions
\author{Abhishek Sinha\thanks{Equal Contribution as first authors. Ordering done alphabetically.} \\
Adobe Systems, Noida\\
{\tt\small abhsinha@adobe.com}
\and
Akilesh B\footnotemark[1] \thanks{Work done when author was at Adobe Systems}\\
Mila, Universit\'e de Montr\'eal\\
{\tt\small akilesh.badrinaaraayanan@umontreal.ca}
\and 
Mausoom Sarkar \\
Adobe Systems, Noida\\
{\tt\small msarkar@adobe.com}
\and
Balaji Krishnamurthy \\
Adobe Systems, Noida\\
{\tt\small kbalaji@adobe.com}
}

\maketitle
\ifwacvfinal\thispagestyle{empty}\fi

%%%%%%%%% ABSTRACT
\begin{abstract}
In this work, we focus on the problem of grounding language by training an agent to follow a set of natural language instructions and navigate to a target object in an environment. The agent receives visual information through raw pixels and a natural language instruction telling what task needs to be achieved and is trained in an end-to-end way. We develop an attention mechanism for multi-modal fusion of visual and textual modalities that allows the agent to learn to complete the task and achieve language grounding. Our experimental results show that our attention mechanism outperforms the existing multi-modal fusion mechanisms proposed for both 2D and 3D environments in order to solve the above-mentioned task in terms of both speed and success rate. We show that the learnt textual representations are semantically meaningful as they follow vector arithmetic in the embedding space. The effectiveness of our attention approach over the contemporary fusion mechanisms is also highlighted from the textual embeddings learnt by the different approaches. We also show that our model generalizes effectively to unseen scenarios and exhibit \textit{zero-shot} generalization capabilities both in 2D and 3D environments. The code for our 2D environment as well as the models that we developed for both 2D and 3D are available at \href{https://github.com/rl-lang-grounding/rl-lang-ground}{https://github.com/rl-lang-grounding/rl-lang-ground}. 
\end{abstract}

%%%%%%%%% BODY TEXT
\section{Introduction}
\label{intro}

Understanding of natural language instructions is an important aspect of an Artificial Intelligence (AI) system. In order to successfully accomplish tasks specified by natural language instructions, an agent has to extract representations of language that are semantically meaningful and ground it in perceptual elements and actions in the environment.

Consider a task in which an agent has to learn to navigate to a target object in an environment. The agent receives visual information through raw pixels and a natural language instruction at the beginning of every episode which specifies the characteristics of the target object. The environment consists of several other objects as well as many obstacles which act as distractors. The agent sees a bird's-eye view of the environment. The challenges that the agent must tackle here are manifold: a) the agent has to develop the capability to recognize various objects, b) have some memory of objects seen in previous states while exploring the environment as the objects may occlude each other and/or may not be present in the agent’s field of view c) decompose the instruction and ground words in visual elements and actions in the environment and d) learn a policy to navigate to the target object by avoiding the obstacles and other non-target objects.  

\begin{wrapfigure}{r}{3cm}
\includegraphics[width=3cm]{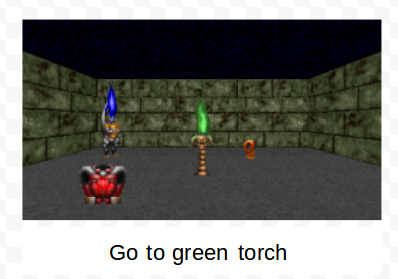}
\includegraphics[width=2cm]{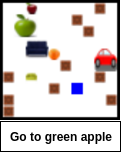}

\caption{The agent should learn to understand the instruction and navigate to the desired locations.}\label{env_1}
\end{wrapfigure} 

We tackle this problem by proposing an end-to-end trainable architecture that creates a combined representation of the image observed by the agent and the instruction it receives. We develop a simple attention mechanism for multi-modal fusion of visual and textual modalities. In order to simulate the above described challenges, we introduce a new 2D environment for an agent to jointly learn visual and textual modalities. We then solve a similar task in 3D vizdoom environment using the same fusion architecture developed in 2D to verify its effectiveness. Finally, in order to enable the reproducibility of our research and foster further research in this direction, we open source the code for our 2D environment as well as the models that we developed for both 2D and 3D. 

\section{Related work}
\label{related_work}

The task of grounding natural language instructions has been well explored by researchers in different domains. From the robotics domain : (\cite{guadarrama2013grounding}, \cite{tellex2011understanding}) focus on grounding verbs in navigational instructions like \textit{go}, \textit{pick up}, \textit{move}, \textit{follow} etc. (\cite{chao2011towards}, \cite{lemaignan2012grounding}) ground various concepts through human-robot interaction. \cite{tellex2011understanding} also focus on manipulation verbs like \textit{put}, \textit{take}. (\cite{artzi2013weakly}, \cite{misra2016tell}) focus on grounding natural language instructions by mapping instructions to action sequences in 2D and 3D environments respectively. Given a natural language instruction, \cite{chen2011learning} attempt to learn a navigation policy that is optimal in a 2D maze-like setting by relying on a semantic parser. \cite{mei2016listen} focus on neural mapping of navigational instructions to action sequences by representing the state of the world using bag-of-words visual features. 

Deep reinforcement learning agents have been previously used to solve tasks in both 2D~\cite{mnih2015human} and 3D~\cite{mnih2016asynchronous} environments. More recently, these approaches have been used for playing first-person shooter games in~\cite{lample2017playing} and~\cite{kempka2016vizdoom}. These works focus on learning optimal policy for different tasks using only visual features, while our work involves the agent receiving natural language instruction in addition to visual state of the environment. 

The authors in~\cite{interactive2018grounded} propose a framework in which an agent learns to navigate and answer questions in 2D maze-like environment (XWORLD). They use a 7x7 2D grid environment with 2-13 worded instructions. We use a larger grid size (10x10) and more complex instructions in which sentence length varies between 3 to 18 for our 2D environment. In addition to having different instances of the same object that differ in their color (for example : green apple, red apple), we also have a size attribute associated with every object which can be either small, medium or large. Further, we also show our approach to work well on 3D environments too.  Another recent work \cite{understanding2018grounded} uses a simple concatenation of visual and textual representations.~\cite{chaplot2017gated} propose a Gated-Attention architecture for task oriented language grounding and evaluate their approach on an environment built over VizDoom~\cite{kempka2016vizdoom}. They do hadamard product between textual and visual representations, whereas we try to create a compact joint representation by stacking attention maps that are obtained by projecting textual features through multiple parallel Fully-Connected (FC) layers and then using each of them to convolve with the visual features. This leads to both reduced memory footprint and faster convergence of the network.

In our work, we present a simple attention based architecture in order to ground natural language instructions in a simulated environment. Our model does not have any prior information of both the visual and textual modalities and is end-to-end trainable without the requirement of external semantic parser or language models as with prior work in robotics domain mentioned previously. In addition, we also show that our network architecture scales to 3D VizDoom environment and outperforms the existing benchmarks, thereby showing the generalizability of our fusion approach to both 2D and 3D environments. 

\section{Problem Description}
\label{prob_desc}
We tackle the problem in which an agent learns to navigate to the target object in a simulated environment. We first perform experiments on our own 2D grid based environment to figure out the correct fusion mechanism of visual and textual features and then apply this mechanism over the VizDoom based 3D environment(~\cite{chaplot2017gated}) to see the extensibility of the method. The agent receives a natural language instruction at the beginning of every episode which specifies the characteristics of the target object. The episode terminates when the agent reaches the target object or the number of time steps exceed the maximum episode length. The environment consists of several other objects as well as many obstacles which act as distractors. The agent sees a bird's-eye view of the environment in which the objects present outside a certain predefined radius are not visible and hence the agent does not have complete knowledge about the environment. The objective of the agent is to learn an optimal policy so that it reaches the correct object before the episode terminates.

\section{Environment}
\label{environment}

\begin{figure}
    \centering
    \begin{minipage}{0.3\textwidth}
        \centering
        \includegraphics[width=0.3\linewidth]{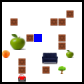} % first figure itself
         \subcaption{Complete view of environment.}
        \label{env:1}
    \end{minipage}
    \begin{minipage}{0.3\textwidth}
        \centering
        \includegraphics[width=0.3\linewidth]{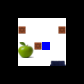} % second figure itself
     \subcaption{Bird's-eye view as seen by agent.}
     \label{egocentric}
    \end{minipage}
    \caption{Example state of environment.}
    \label{env}
\end{figure}
We perform experiments on two different environments :

a) A 2D environment created by us and\\
b) Vizdoom based 3D environment.

In our 2-D grid based environment the agent interacts with the environment and performs one of the four actions: \textit{up}, \textit{down}, \textit{left} and \textit{right}. Each scenario in the environment consists of an agent, a list of objects and a list of obstacles (walls) as shown in figure~\ref{env:1}. Every object and obstacle has a set of attributes associated with it like color, size etc. The environment is customizable as the grid size, number of objects, type of objects, number of obstacles along with their corresponding attributes can be specified in a configurable specification file which is in JSON format (refer supplementary section for an example configuration file). The agent perceives the environment through raw RGB pixels with a bird's-eye view, in which we blacken out the region outside certain predefined radius centered at the agent(figure~\ref{egocentric}). At the beginning of every episode, the agent receives a natural language instruction and obtains a positive reward (+1) on successful completion of the task. The task is considered successful if the agent is able to navigate to the target object correctly before the episode ends . The agent gets a negative reward (-1) whenever it hits a wall. The agent also receives a small negative (-0.5) reward if it reaches any non-target object. In every episode, the environment is reset randomly, i.e., the agent, the target object, non-target objects are placed at random locations in the grid. The instruction is generated based on the initial configuration of the environment. Some example instructions are given below:-

\begin{itemize}

\item \textit{Go to car} or \textit{Car is your target(or destination)}. 

\item \textit{Go to green apple} or \textit{Green apple is your target(or destination)}. 

\item \textit{Go to medium blue sofa} or \textit{Medium blue sofa is your target(or destination)}. 

\item \textit{Go to south of blue bus}. 

\item \textit{Go to bottom right corner}. 

\item \textit{There is a orange chair}. \textit{Go to it}. 
%%% Defer the dereferencing part for now
%The agent is required to infer here that 'it' refers to orange chair and then navigate to it accordingly.

\item \textit{There is a small blue bag}. \textit{Go to it}. 

\item \textit{There are multiple green tree}. \textit{Go to smaller(or larger) one}.  

\item \textit{There is a small yellow banana and a medium black television. Go to former(or latter)}. 

\item \textit{If small yellow flower is present then go to small orange cat else go to medium black bear}. 

%% Add other sentences as we get results.  
\end{itemize}

The complete set of objects and instructions are provided in the supplementary material. 

\section{Proposed Approach}
\label{proposed_approach}

\begin{figure}
\includegraphics[width=\columnwidth]{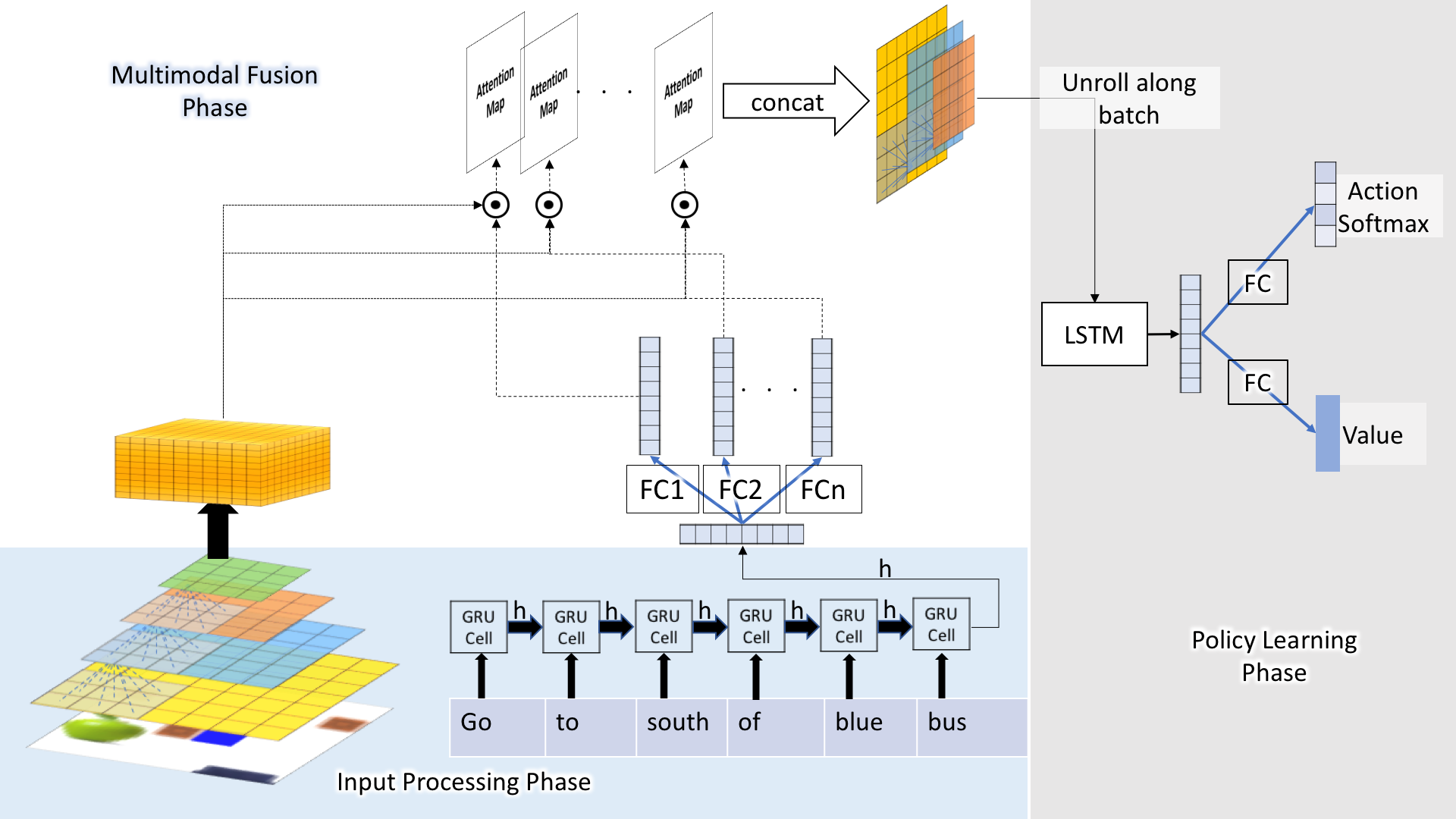}
\caption{Our network architecture consisting of all three phases.}\label{network_1}
\end{figure}

\begin{figure}
    \centering
    \begin{minipage}{0.55\textwidth}
        \centering
        \includegraphics[width=0.75\linewidth]{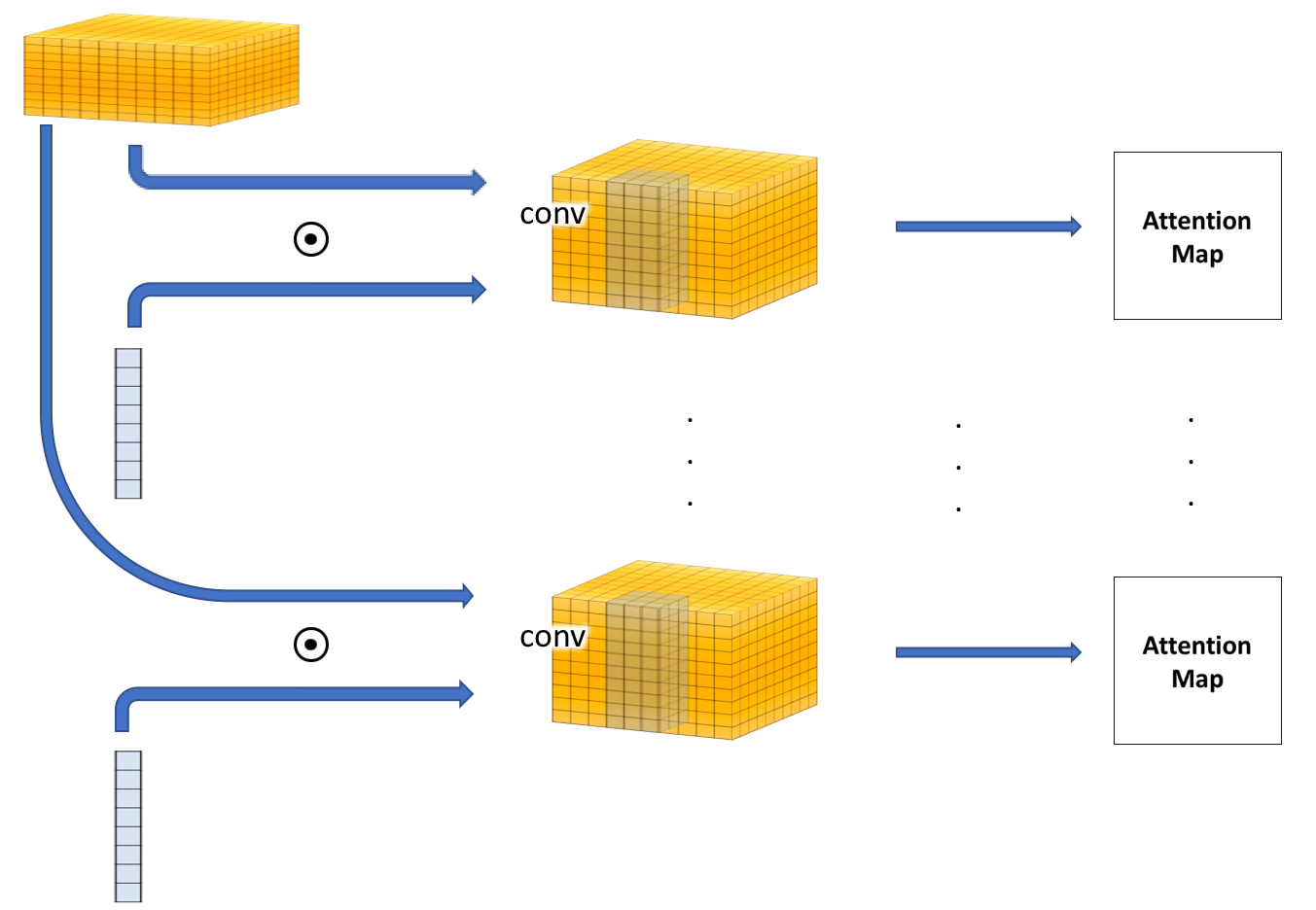} % first figure itself
         \caption{Multimodal fusion phase: each vector obtained from FC layer is used as 1x1 filter to perform convolution over visual representation}
        \label{env:11}
    \end{minipage}
\end{figure}
The proposed model can be divided into three phases a) input processing phase b) multi-modal fusion phase and c) policy learning phase, as illustrated in figure~\ref{network_1}

\subsection{Input processing phase}
\label{ipphase}
The agent at each time step ($t$) receives a RGB image of the environment which is a bird's-eye view in case of 2D and ego-centric view in case of 3D ($E_{t}$) along with an instruction ($I$) at the start of the episode. The image $E_{t}$ is passed through several layers of convolutional neural network (CNN)  to obtain a good representation of the image. The final representation of the image obtained in this phase is $R_{E}$ and is of dimension $W \times H \times D$, where $W$, $H$ are width and height of each feature map and $D$ represents the number of feature maps. (Refer section ~\ref{impl_details} for exact details of network hyper-parameters). Every word in the instruction $I$ is converted into an one hot vector and then concatenated before passing it through the GRU. For the 2D environment, the maximum number of words present in the instruction is 18 and the total number of possible unique words (i.e. \textit{vocabulary size}) is 72. The complete list of these words are provided in supplementary section. The concatenated one hot vectors are passed through a GRU and the output of the final time step is used as a representation of the entire sentence. The output of the GRU is then passed through multiple parallel (say $n$) FC layers. This results in generation of vectors $V_{1}$, $V_{2}$ \ldots ,$V_{n}$.

\subsection{Multimodal fusion phase}
\label{mmfusion}
We propose a simple attention method for the multimodal fusion of $R_{E}$ and $V_{j}$ where $j$ $\in$ \{1,2, \ldots ,n\}. Each $V_{j}$ is used as a $1 \times 1$ filter to perform 2D convolution over feature volume $R_{E}$ which is of dimension $W  \times H \times D$ resulting in an attention map of dimension $W \times H  \times 1$. All these attention maps are then concatenated to result in a tensor of dimension $W \times H \times n$ (say $M_{attn}$ and the corresponding model $A3C_{attn}$). In our experiments we try with $n$ $\in$ \{1,5,10\} (refer section~\ref{results} for comparison of performance). $M_{attn}$ is then fed as input to the policy learning phase. Our motivation behind this fusion mechanism was to discard unnecessary information and obtain a joint minimalistic representation ($M_{attn}$) of both modalities which is in-line with the concept of consciousness of an AI agent as highlighted by ~\cite{bengio2017consciousness}. We also consider a small variant of this fusion mechanism where we concatenate one of the attention maps with the visual representation resulting in a tensor of dimension $W \times H \times (D+1)$ (say $M_{netattn}$ and the corresponding model $A3C_{netattn}$). We also compare our method with previously proposed multimodal fusion methods: a) Gated Attention unit~\cite{chaplot2017gated} (say $A3C_{hadamard}$) b) Concatenation unit~\cite{understanding2018grounded} (say $A3C_{concat}$). 

\subsection{Policy learning phase}
This phase receives as input, the output of the multimodal fusion phase. We adopt a reinforcement learning approach using the Asynchronous Advantage Actor-Critic (A3C)~\cite{mnih2016asynchronous} algorithm. In A3C, there is a global network and multiple local worker agents each having their own network parameters. Each of these worker agents interacts with it's own copy of environment simultaneously and gradients of these worker agents are used to update the global network. Refer section~\ref{impl_details} for exact details of our architecture.

\section{Experimental Setup}
\label{exp_setup}

For the 2D environment, we fix the grid size to be 10x10 with 63 objects and 8 obstacles. At the start of every episode, we randomly select between 3 to 6 objects from the set of 63 objects and a single instruction is randomly selected from a set of feasible instructions. For our experiments on 3D environment, we use the exact same environment used by the authors in~\cite{chaplot2017gated} which has been open-sourced~\cite{chaplot2017code}.
We use \textit{success rate} of the agent and \textit{mean reward} it obtains as the evaluation metrics. The \textit{success rate} is the number of times the agent reaches the target object successfully before the episode terminates and \textit{mean reward} is the average reward that the agent obtains across all episodes. We consider two modes of evaluation: 
\par 
1) \textbf{Unseen scenario generalization(US):} At test time, the agent is evaluated in unseen environment scenarios with instructions in the train set. The scenario consists of combination of objects placed at random locations not seen before by the agent at train time.
\par
2) \textbf{Zero-shot generalization(ZS):} At test time, the agent is evaluated with unseen instructions that consists of new combinations of attribute-object pairs never seen during training. The environment scenario is also unseen in this case.

\subsection{Implementation details}
\label{impl_details}
The implementation details for the case of 2D environments are as follows :\\
The input to the input processing phase neural network is a RGB image of size 84x84x3 and an instruction. The input image is processed with a CNN that has four convolution layers. The first layer convolves the image with 32 filters of 5x5 kernel size with stride 2, followed by another 32 filters of 5x5 kernel size with stride 2. This is then followed by 64 filters of 4x4 kernel size with stride 1 and finally by another 64 filters of 3x3 kernel size with stride 2. The input instruction is encoded through a GRU of size 16. The encoded instruction is then passed through a FC layer of size 64. The visual mode is then combined with textual mode through our attention mechanism. We also evaluate various multi-modal fusion mechanisms like: a) simple concatenation, b) our attention mechanism and its variants, c) Gated-Attention unit (Hadamard product). In case of our attention mechanism, the multiple attention maps are concatenated and passed through two other convolutional layers (each having 64 filters of 3x3 kernel size with stride 1) before passing it to the LSTM layer. All our experiments are performed with A3C algorithm. Our policy learning phase has a LSTM layer of size 32, followed by fully connected layer of size 4 to estimate the policy function as well as fully connected layer of size 1 to predict value function. All the convolutional layers and FC layers have PReLU activations~\cite{he2015delving}. We observed during our experimentation the importance of not suppressing the negative gradients, as the same architecture in which all convolutional and FC layers have ReLU activations~\cite{nair2010rectified} performed poorly on our evaluation scenarios (Refer section~\ref{results} for this comparison). The A3C algorithm was trained using Adam optimizer~\cite{kingma2014adam} with an annealing learning rate schedule starting with 0.0001 and reducing by a fraction of 0.9 after every 10000 steps. For each experiment, we run 32 parallel threads and we use a discount factor of 0.99 for calculating expected rewards. As described in~\cite{mnih2016asynchronous} we use entropy regularization for improved exploration. Further, in order to reduce the variance of the policy gradient updates, we use the Generalized Advantage Estimator~\cite{schulman2015high}.\\  

For the 3D environment case, we use the same architecture and hyperparameters as mentioned by the authors in~\cite{chaplot2017code}, and train from scratch by replacing their hadamard product based fusion mechanism with our attention mechanism.

\section{Results and Discussions}
\label{results}
Table~\ref{attn_comp} depicts the performance of our attention mechanism ($A3C_{attn}$) as described in section~\ref{mmfusion} for various $n$ values for 2D environment.

\begin{table}[h]
\caption{The mean reward and succcess rate of our model for different $n$ values on unseen scenario generalization(US) and ZS instructions for 2D environment}

\begin{tabular}{|p{3cm}|p{1.2cm}|p{1cm}|p{1cm}|} \hline 
Model &mean reward(US) & mean reward (ZS) & success rate  \\ \hline 
$A3C_{attn}$ ($n=1$) & 0.87  & 0.73 & 0.86\\ \hline
$A3C_{attn}$ ($n=5$) & \textbf{0.95} & \textbf{0.8} & \textbf{0.92}\\ \hline
$A3C_{attn}$ ($n=10$) & 0.94& 0.75 & 0.88\\  
\hline
\end{tabular}
\label{attn_comp}
\end{table}

The \textit{success rate} and \textit{mean reward} values are averaged over 100 episodes. We observe from the table that $n=5$ achieves best \textit{mean reward} of $0.95$. We also found that when $A3C_{attn}$ ($n=5$) is evaluated under \textit{zero-shot} generalization settings, it obtains a \textit{mean reward} of $0.8$. For \textit{zero-shot} evaluation, we measure the agent's success rate on 19 instructions that were held out during training and these instructions consist of new combinations of attribute-object pairs not seen before by the agent during train phase (refer supplementary section for these instructions).
Table~\ref{base_comp} portrays the performance of $A3C_{attn}$ with $n=5$ with a variant of our attention method ($A3C_{netattn}$) as well as with other previously proposed multimodal fusion mechanism methods ($A3C_{hadamard}$ and $A3C_{concat}$) as described in
section~\ref{mmfusion}

\begin{figure}
    \centering
    \begin{minipage}{0.4\textwidth}
        \centering
        \includegraphics[width=\linewidth]{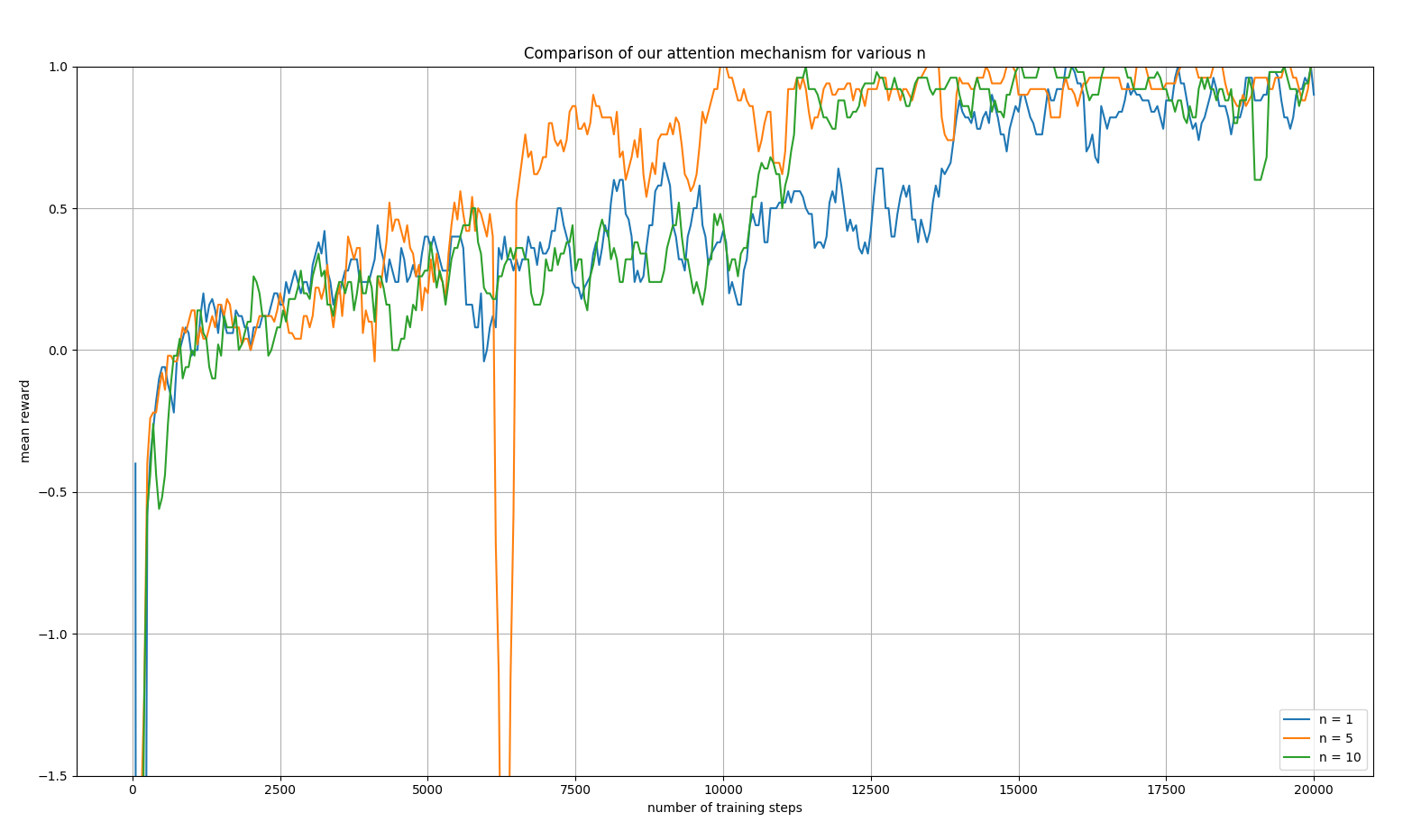} % first figure itself
         \subcaption{Performance of $A3C_{attn}$ ($n=1,5,10$).}
        \label{results:1}
    \end{minipage}
    \begin{minipage}{0.4\textwidth}
        \centering
        \includegraphics[width=\linewidth]{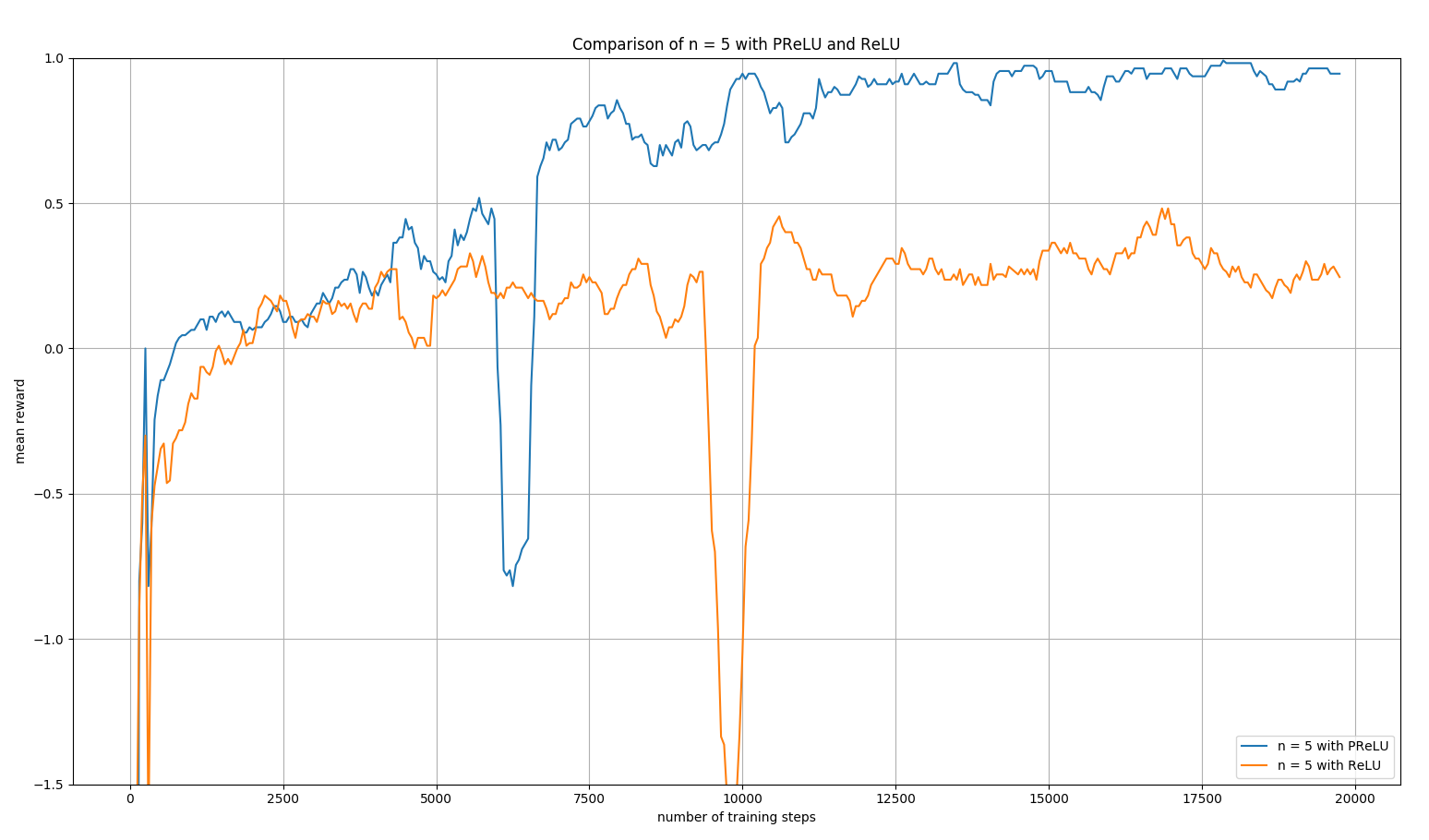} % second figure itself
         \subcaption{Performance of $A3C_{attn}$ ($n=5$) with PReLU vs ReLU.}
     \label{results:2}
    \end{minipage}
    \begin{minipage}{0.4\textwidth}
        \centering
        \includegraphics[width=\linewidth]
        {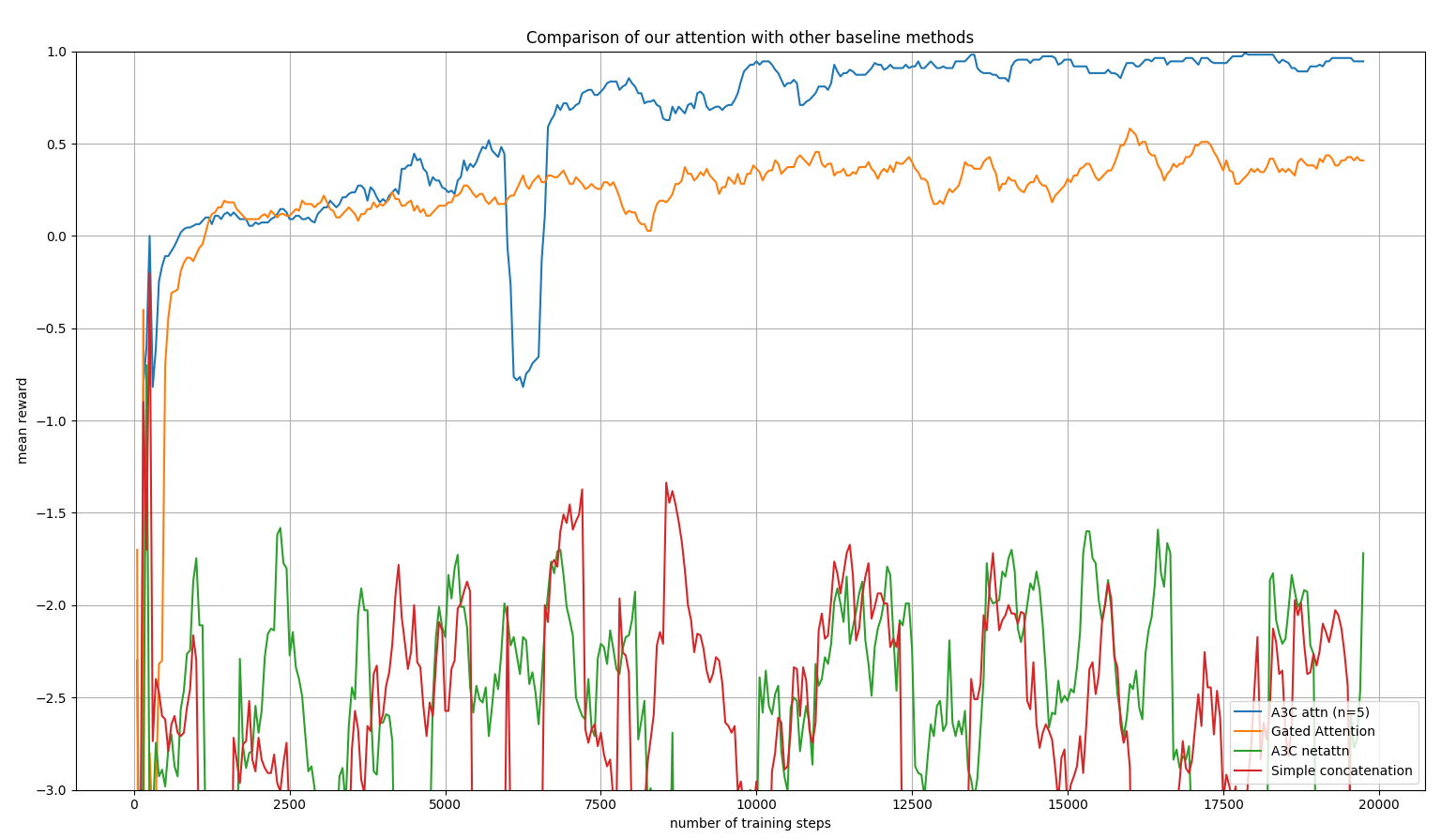} % second figure itself
        \subcaption{Performance of $A3C_{attn}$ ($n=5$) with other baseline methods.}
     \label{results:3}
    \end{minipage}
    \caption{Performance analysis of our attention mechanism and comparison with other baseline methods.}
    \label{results}
\end{figure}

\begin{table}[h]
\centering
\caption{Comparison of our model with other baseline methods on unseen scenario generalization for 2D environment}
\begin{tabular}{|c|c|c|} \hline 
Model & mean reward & success rate  \\ \hline 
$A3C_{attn}$ ($n=5$) & \textbf{0.95} & \textbf{0.92}\\ \hline
$A3C_{netattn}$ & -2.8 & 0.1\\ \hline
$A3C_{hadamard}$ & 0.4  & 0.5\\ \hline
$A3C_{concat}$ & -2.9 & 0.1\\  
\hline
\end{tabular}
\label{base_comp}
\end{table}
Figure ~\ref{results:1} shows the \textit{mean reward} obtained by our $A3C_{attn}$ model for different $n$ values as the training progresses. We observe from the graph that $n=5$ converges faster to higher reward values. Figure~\ref{results:2} shows the comparison of our best performing $A3C_{attn}$ model in which all convolutional and FC layers have PReLU activations against the case in which all convolutional and FC layers have ReLU activations. It is evident from the graph the performance gain obtained just by not suppressing negative gradients using PReLU and accentuates the importance of choosing appropriate activation function.
From figure~\ref{results:3}, it is apparent that our $A3C_{attn}$ with $n=5$ converges faster and performs significantly better than other methods.
 
\begin{figure}
    \centering
    \begin{minipage}{0.4\textwidth}
        \centering
        \includegraphics[width=\linewidth]{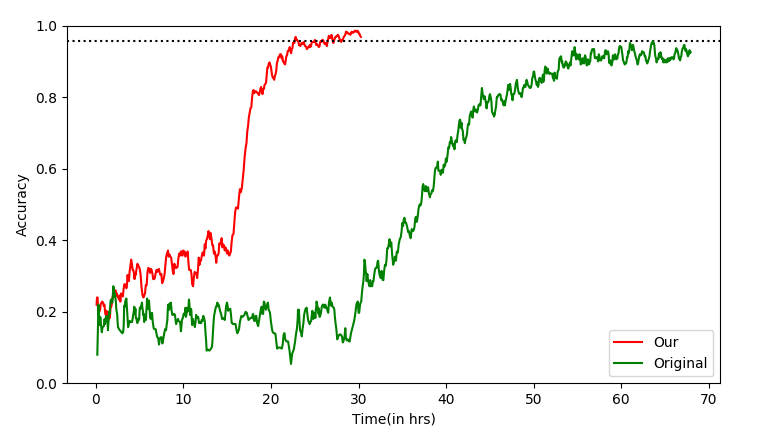} % first figure itself
         \subcaption{Easy difficulty}
        \label{results3d:1}
    \end{minipage}
    \begin{minipage}{0.4\textwidth}
        \centering
        \includegraphics[width=\linewidth]{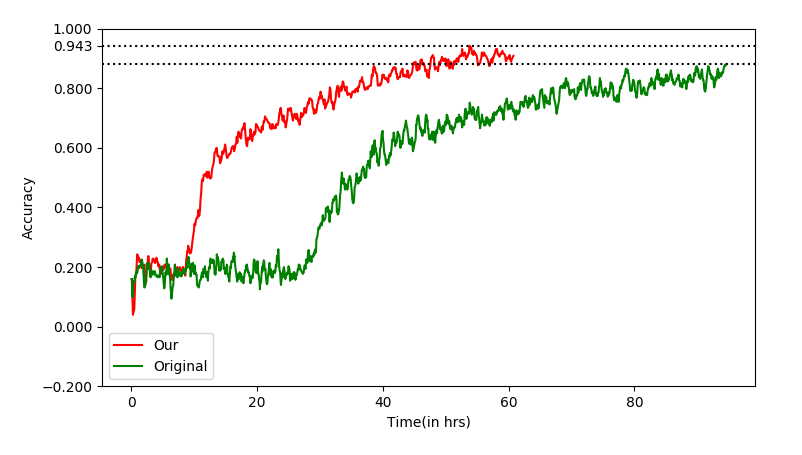} % second figure itself
         \subcaption{Medium difficulty}
     \label{results3d:2}
    \end{minipage}
    \begin{minipage}{0.4\textwidth}
        \centering
        \includegraphics[width=\linewidth]{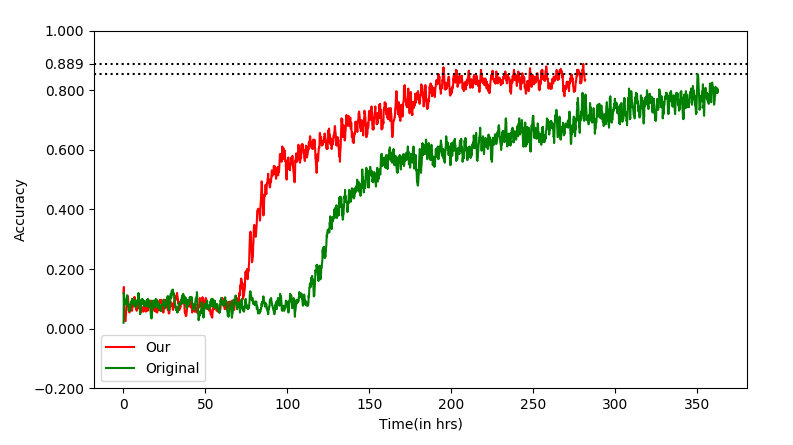} % second figure itself
         \subcaption{Hard difficulty}
     \label{results3d:3}
    \end{minipage}

    \caption{Comparison of our attention based fusion mechanism with the fusion mechanism by \cite{chaplot2017gated} for 3d environment.}
    \label{results_3d}
\end{figure}

Table ~\ref{base_comp_3D} compares the performance of our attention mechanism in 3D environment with the other baselines for three different levels of difficulty : \textit{Easy}, \textit{Medium} and \textit{Hard}. The details of these difficulty levels are elucidated in ~\cite{chaplot2017gated}. Figure ~\ref{results_3d} compares our attention mechanism with the hadamard-product based attention~\cite{chaplot2017gated} during the progress of training for different difficulty levels. As can be seen from both the table as well as the figure, our model not only converges much faster than the actual baseline but also improves the performance in terms of success rate(both for zero shot instructions and unseen scenario generalization) for 3D environments too. The other proposed fusion mechanisms either work only in the 2D or 3D environment but not in both, whereas our attention based fusion method performs equally good in both the environments.

\begin{table}[h]
\centering
\caption{Comparison of our model with other baseline methods for 3D environment with different difficulties}
\begin{tabular}{|c|c|c|c|c|c|c} \hline 
Model & \multicolumn{2}{|c|}{Easy} & \multicolumn{2}{|c|}{Medium}  & \multicolumn{2}{|c|}{Hard}  \\ \hline
& US & ZS & US & ZS & US &ZS \\ \hline
$A3C_{attn}$ ($n=5$) & 1.0 & \textbf{0.95} & \textbf{0.92} & \textbf{0.9} & \textbf{0.86} & \textbf{0.83} \\ \hline

$A3C_{hadamard}$ & 1.0 & 0.81 & 0.89 & 0.75 & 0.83 & 0.73 \\ \hline

$A3C_{concat}$ & 1.0 & 0.80 & 0.80 & 0.54 & 0.24 & 0.12 \\

\hline
\end{tabular}
\label{base_comp_3D}
\end{table}

\begin{figure*}
    \centering
    \begin{minipage}{0.45\textwidth}
        \centering
        \includegraphics[width=0.45\linewidth]{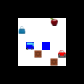} % first figure itself
        \subcaption{Bird's-eye view as seen by agent.}
           	\label{attn1_env}
    \end{minipage}
    \begin{minipage}{0.45\textwidth}
       \centering
        \includegraphics[width=0.45\linewidth]{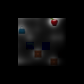} % second figure itself
        \subcaption{Attention map 1}
         \label{attn1_weights1}
    \end{minipage}
    \begin{minipage}{0.45\textwidth}
       \centering
        \includegraphics[width=0.45\linewidth]{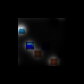} % second figure itself
        \subcaption{Attention map 2}
        \label{attn1_weights2}
    \end{minipage}
    \begin{minipage}{0.45\textwidth}
       \centering
        \includegraphics[width=0.45\linewidth]{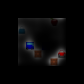} % second figure itself
         \subcaption{Attention map 3}
        \label{attn1_weights3}
    \end{minipage} 
    \caption{Visualization of attention maps for sentence: "Go to small red car."}
    \label{attn1_weights}
\end{figure*}

We also visualize the attention maps for our best performing $A3C_{attn}$ model over 2D environment. Figure~\ref{attn1_weights} shows the visualization of attention maps for the case when the sentence is "Go to small red car." As shown in figures~\ref{attn1_weights1},~\ref{attn1_weights2} and~\ref{attn1_weights3}, different attention maps focus on different regions in the environment and the agent uses a combination of these in order to learn the policy and successfully navigate to the target object while
avoiding incorrect objects and obstacles. Attention map~\ref{attn1_weights3} seems to highlight the possible target objects while attention maps~\ref{attn1_weights1} and~\ref{attn1_weights2} appear to be focusing on non-target objects.

\begin{figure}
\centering
    \begin{minipage}{0.55\textwidth}
         \centering
        \includegraphics[width=0.55\linewidth]{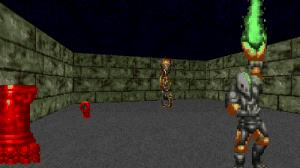} % first figure itself
        \subcaption{Egocentric view as seen by agent.}
           	\label{attn1_env_supp}
    \end{minipage}
    
    \begin{minipage}{0.55\textwidth}
      \centering
        \includegraphics[width=0.55\linewidth]{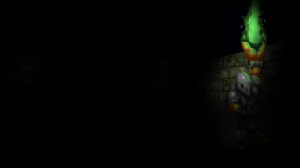} % second figure itself
        \subcaption{Attention map 1}
         \label{attn1_weights1_3D}
    \end{minipage}
    \begin{minipage}{0.55\textwidth}
       \centering
        \includegraphics[width=0.55\linewidth]{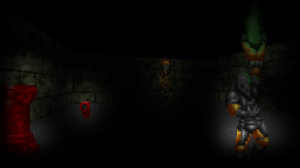} % second figure itself
        \subcaption{Attention map 2}
        \label{attn1_weights2_3D}
    \end{minipage}
    \caption{Visualization of attention maps for sentence: "Go to tall green torch"}
    \label{attn1_weights_3D}
\end{figure}

In a similar manner, we also show some of the attention maps of our model trained over the 3D environment in figure~\ref{attn1_weights_3D}.

In order to understand the quality of instruction embedding learnt by the GRU, we do a two-dimensional Principal Component Analysis (PCA) projection of the original 16 dimension representation obtained when the input instruction is encoded through a GRU of size 16. As shown in figure~\ref{pca_1}, \textit{vector("Go to green apple") - vector("Go to apple")} is parallel to \textit{vector("Go to green sofa") - vector("Go to sofa")}. Similarly \textit{vector("Go to blue bag") - vector("Go to bag")} is parallel to \textit{vector("Go to blue bus") - vector("Go to bus")}. We also observe a similar pattern with instructions that include size attribute in addition to color attribute. From
figure~\ref{pca_2}, we discern that \textit{vector("Go to medium blue bag") - vector("Go to blue bag")} is parallel to \textit{vector("Go to medium green tree") - vector("Go to green tree")}.  These observations point to our model's ability to organize various concepts and learn the relationships between them implicitly.
% In figure ~\ref{pca_3}, we notice similar relationship with instructions that specify direction.

\begin{figure}
    \centering
    \begin{minipage}{0.6\textwidth}
        % \centering
        \includegraphics[width=0.65\linewidth]{images/pca1.png} % first figure itself
        \subcaption{Instruction type - Go to [object] or Go to [color][object]}
     \label{pca_1}
    \end{minipage}
    \begin{minipage}{0.6\textwidth}
        % \centering
        \includegraphics[width=0.65\linewidth]{images/PCA_size.png} % second figure itself
        \subcaption{Instruction type - Go to [color][object] or Go to [size][color][object]}
     \label{pca_2}
    \end{minipage}
    % \begin{minipage}{0.65\textwidth}
    %     \centering
    %     \includegraphics[width=0.65\linewidth]{images/PCA_direction.png} % second figure itself
    %     \subcaption{Instruction type- Go to [dir.] of [color][object] or Go to [size][color][object]}
    %  \label{pca_3}
    % \end{minipage}
    \caption{Two-dimensional PCA projection of instruction embedding over the 2D environment.}
    \label{pca_plot}
\end{figure}

\begin{figure}
%   \centering
    \begin{minipage}{0.65\textwidth}
    % \centering
        \includegraphics[width=0.65\linewidth]{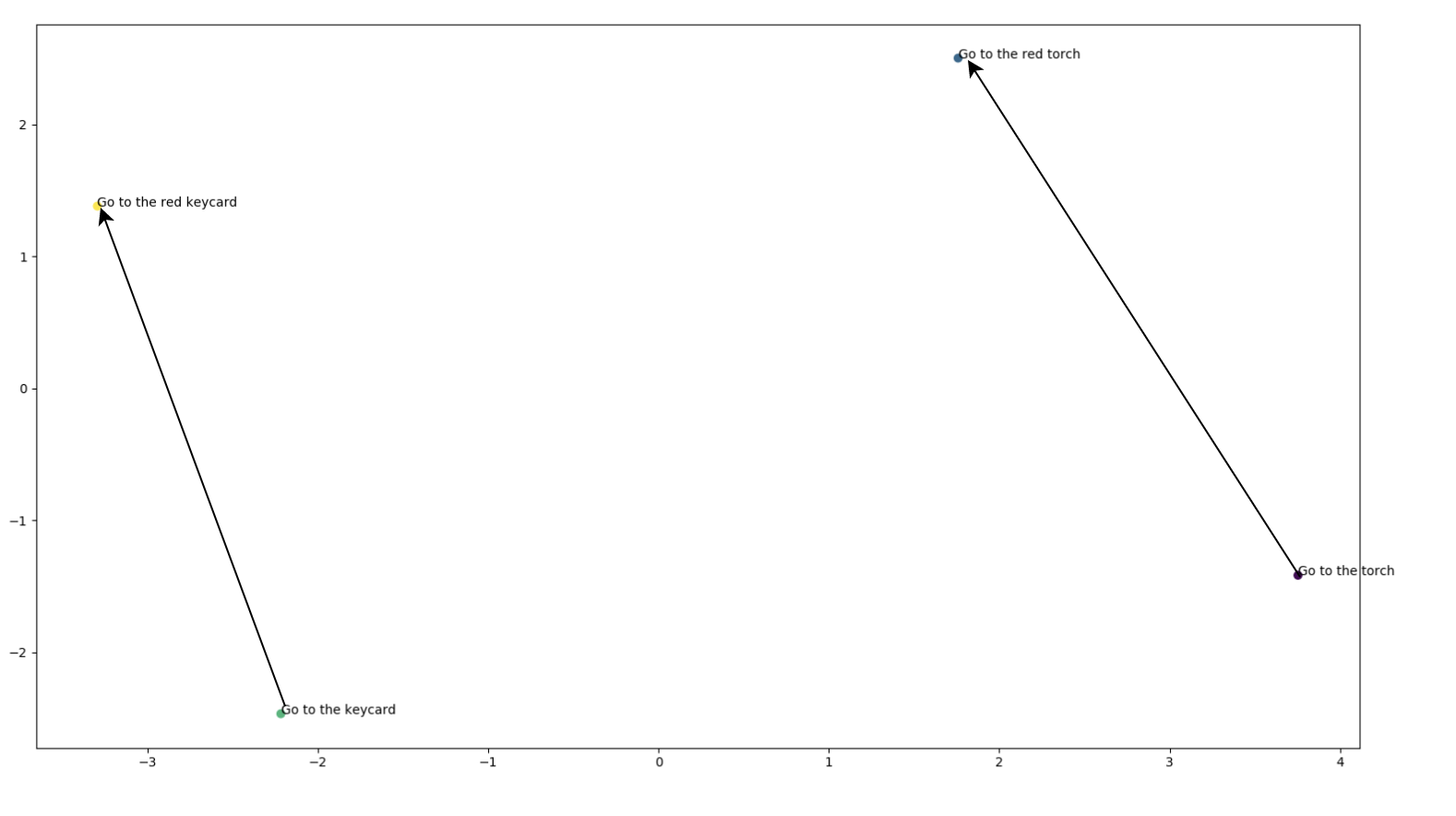} % first figure itself
        \subcaption{Embedding vectors obtained by $A3C_{attn}$}
        \label{pca_1_3D}
    \end{minipage}
    \begin{minipage}{0.65\textwidth}
    % \centering
        \includegraphics[width=0.65\linewidth]{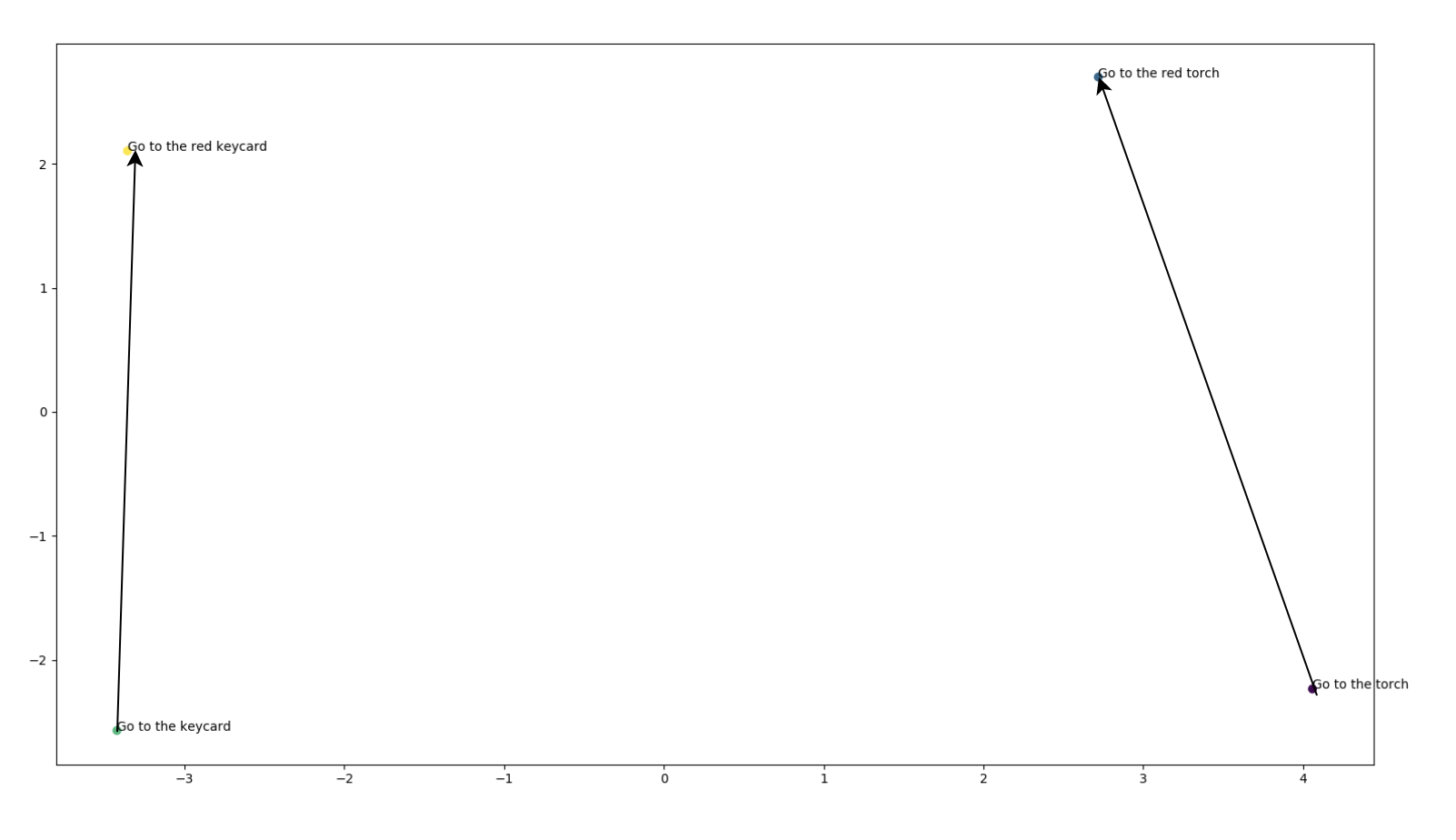} % second figure itself
        \subcaption{Embedding vectors obtained by $A3C_{hadamard}$}
     \label{pca_2_3D}
    \end{minipage}
    
    \caption{Comparison of two-dimensional PCA projection of instruction embedding for 3D environment}
    \label{pca_plot_3D}
\end{figure}

We also generate the embedding vector for the case of 3D environment and compare the vectors obtained by our fusion mechanism($A3C_{attn}$) with the baseline($A3C_{hadamard}$) approach in figure ~\ref{pca_plot_3D}. As can be seen from the figures the \textit{vector("Go to the red torch") - vector("Go to the torch")} is nearly parallel to \textit{vector("Go to the red keycard") - vector("Go to the keycard")} for our approach whereas this is not true for the baseline approach. This further validates the higher zero-shot success rate achieved by our approach.

Additionally, for the experiments on 2D, we also find that these instruction embeddings follow vector arithmetic. In order to validate this, we obtain the embeddings for the
instruction \textit{"Go to small green tree"} as follows: \textit{vector("Go to tree") + (vector("Go to green apple") - vector("Go to apple")) + vector("Go to small red car") - vector("Go to red car")}. Using this, the agent learns to navigate successfully to the small green tree. Again, the embedding calculated by \textit{vector("There are multiple green Tree. Go to smaller one") - (vector("Go to small blue Bag") - vector("Go to blue Bag")) + (vector("Go to medium blue Bag") - vector("Go to blue Bag"))}, makes the agent go to medium green tree even when small green tree is present in the environment(refer to the supplementary material for depiction of this behaviour). The agent even responds to sentences such as "Go to small Apple" formed by \textit{vector("Go to Apple") + vector("Go to small green Sofa") - vector("Go to green Sofa")} even though the agent has never seen sentences of the form \textit{"Go to [size][object]"}.    
\subsection{Language translation as a by-product of grounding}
Language grounding eventually means capturing the essence of words which consequently should enable an agent to translate the same concept across different languages. For e.g if an agent knows that the concept 'apple' in multiple languages represent the same object, then given any word which represent the concept 'apple' in any of its known languages, the agent should be able to translate it to other known languages. In order to examine if our agent is grounded well enough to do such translations, we train the agent by giving it instructions in both English and
French on a subset of vocabulary. At test time, to assess the instruction embedding, we ask the agent to translate English instructions to French and vice-versa. We found that the agent manages to translate \textbf{85\%} of the instructions. We say an agent successfully translates the instruction if the translated sentence and the ground truth translation match word by word entirely. Some of the translation achieved by the agent are :-

\begin{itemize}
\item Input - Go to small red car. \\
	  Output - Aller a la petite voiture rouge.

\item Input - Aller a la chaise vert. \\
	  Output - Go to green chair.	

\end{itemize}

To achieve the above task, we attach a decoder branch consisting of two separate decoders, one for each English and French, on top of the instruction embedding obtained through GRU. The encoder is kept the same for both the languages. The task of the decoders during the training phase is to reconstruct the original instruction as it is.  Based on whether the natural language instruction is English or French, reconstruction loss through the corresponding decoder is back-propagated to modify the weights of the GRU. However during the test phase the decoder corresponding to each language is switched. Therefore the English decoder is activated while giving French instructions and the French decoder is activated while giving English instructions. It is important to note that during training,  the agent receives instructions in either of the two languages randomly. So the agent does not have a mapping between the two version of instructions. It can only establish the relation between them because it gets a positive reward only by completing the task specified by the instruction. Thus, the agent manages to do the translation in a completely unsupervised way, without being explicitly given a parallel corpora.\linebreak

% It is important to note that the translation was done in a completely unsupervised way, without explicitly giving parallel corpora to the agent.\linebreak

% During training the agent receives either of them randomly. So the agent does not have a mapping between the two version of instructions. It can only establish the relation between them because it gets a positive reward only by completing the specific task. Therefore the mapping is established by mapping each version(English and French) to the common task in the environment.

Thus, all these experiments prove that the agent has learnt to correctly associate the words with their true sense or meaning entirely on its own. In all these experiments, the agent's trajectory as it navigates to the target object are stored in the form of GIFs and are available at  \href{https://github.com/rl-lang-grounding/rl-lang-ground}{https://github.com/rl-lang-grounding/rl-lang-ground}. 

\section{Conclusion and Future Work}
\label{conclusion}
In the paper we presented an attention based simple architecture to achieve grounding of natural language sentences via reinforcement learning. We validate the generalizability of our method by showing faster convergence as well as improved success rate, both for 2D as well as 3D environments,  in comparison to other baselines.
We consistently observed in our experiments that retaining just the representation obtained after multimodal fusion phase (i.e. multiple attention maps) and discarding the visual features helps the agent achieve its goals. Also visualization of the attention maps reveals that they contain enough information needed for the agent to find the optimal policy. Through vector arithmetic, we also show that the embeddings learnt by the agent indeed make sense. In order to encourage the research in this direction we have also open sourced our environment as well as the code and models developed.

Our 2D environment is capable of supporting rich set of natural language instructions and it's flexible as it is easy to add new objects in the environment through the JSON file. Moreover, one can also add new set of instructions via minimal changes in the code. As a future work, we would like to increase the complexity of both the types of sentences generated as well as the environment dynamics by introducing moving objects.

{\small
\bibliographystyle{ieee}
\bibliography{egpaper_final}
}

\end{document}